\documentclass[letterpaper, 10 pt, conference]{ieeeconf}  
\usepackage{times}
\IEEEoverridecommandlockouts                             

\usepackage{bbold}

\usepackage{multicol}
\usepackage[bookmarks=true]{hyperref}
\usepackage{graphicx}
\usepackage{amsfonts}
\usepackage{amsmath,amssymb} 
\usepackage{esvect}
\usepackage{algorithm}
\usepackage[noend]{algpseudocode}
\usepackage{multirow}
\usepackage{subfig}
\usepackage{caption}
\usepackage{soul} 
\usepackage{comment} 
\usepackage{balance} 
\usepackage{bchart}
\usepackage{newfloat}
\usepackage{bbm}
\usepackage{pgfplots}
\usepackage{gensymb}
\usepackage{siunitx}
\usepackage{bbold}

\pgfplotsset{width=9cm,compat=1.8}

\DeclareFloatingEnvironment[
   fileext=loc
]{chart}

\usepackage[colorinlistoftodos,prependcaption,textsize=tiny]{todonotes}

\usepackage[normalem]{ulem}

\begin{document}

\title{Learning a Group-Aware Policy for Robot Navigation}

\author{Kapil Katyal$^{*1,2}$, Yuxiang Gao$^{*2}$, Jared Markowitz$^{1}$,\\ Sara Pohland$^{3}$, Corban Rivera$^{1}$, I-Jeng Wang$^{1}$, Chien-Ming Huang$^{2}$
\thanks{$^{*}$Both authors contributed equally to this work.
        {\tt\small }}%
\thanks{$^{1}$Johns Hopkins University Applied Physics Lab, Laurel, MD, USA.
        {\tt\small }}%
\thanks{$^{2}$Dept. of Comp. Sci., Johns Hopkins University, Baltimore, MD, USA.
        }%
\thanks{$^{3}$Dept. of EECS, UC Berkeley, Berkeley, CA, USA.
        }%
}

\maketitle

\begin{abstract}
Human-aware robot navigation promises a range of applications in which mobile robots bring versatile assistance to people in common human environments. 
While prior research has mostly focused on modeling pedestrians as independent, intentional individuals, people move in groups; consequently, it is imperative for mobile robots to respect human groups when navigating around people. 
This paper explores learning group-aware navigation policies based on dynamic group formation using deep reinforcement learning. 
Through simulation experiments, we show that group-aware policies, compared to baseline policies that neglect human groups, achieve greater robot navigation performance (e.g., fewer collisions), minimize violation of social norms and discomfort, and reduce the robot's movement impact on pedestrians. 
Our results contribute to the development of social navigation and the integration of mobile robots into human environments. 

\end{abstract}

\section{Introduction}
Mobile robots that are capable of navigating crowded human environments in a safe, efficient, and socially appropriate manner hold promise in bringing practical robotic assistance to a range of applications, including security patrol, emergency response, and parcel delivery. 
An increasing body of research has focused on the challenging quest to enable human-aware robot navigation, accounting for human movements that are fast, dynamic, and follow delicate social norms~\cite{charalampous2017recent,pandey2017mobile,kruse2013human,rios2015proxemics}. 
For example, prior research has treated humans as dynamic obstacles to avoid collisions (e.g.,~\cite{fox1997dynamic}), investigated strategies to avoid getting stuck in human crowds (e.g.,~\cite{trautman2015robot}), and explored how to model social norms to allow for socially appropriate robot navigation (e.g.,~\cite{robicquet2016learning,bera2017sociosense,mead2017autonomous}). 
However, prior works have mainly treated people as individual, independent entities in robot navigation.

The majority of people, however, walk in groups~\cite{costa2010interpersonal,aveni1977not}; an empirical study showed that up to 70\% of pedestrians in a commercial environment walked in groups~\cite{moussaid2010walking}. 
Consequently, it is imperative that a mobile robot respects human grouping (e.g., not to intersect through a social group) during its navigation in a human environment. 
In particular, we consider the problem of a robot interacting with dynamic human groups---people walking together in groups---rather than standing groups that are commonly seen in social events (e.g.,~\cite{okal2016learning}). 
While substantial efforts have been made to model dynamic groups (e.g.,~\cite{moussaid2010walking,yucel2013deciphering,bisagno2018group}), 
how mobile robots should navigate effectively and appropriately around dynamic human groups is under-explored.  

\begin{figure}[t] 
	\centering
    \includegraphics[width=1.0\columnwidth]{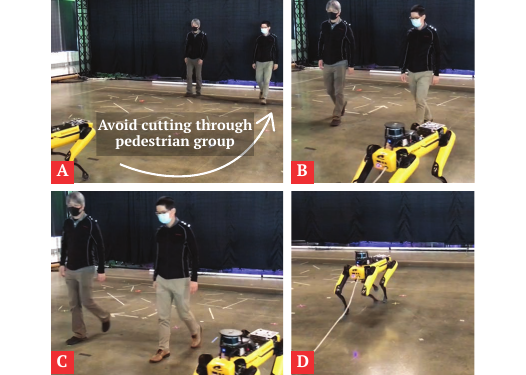}
    \caption{The objective of this work is to learn a navigation policy that allows the robot to safely reach its goal while minimizing impact to individuals and groups of pedestrians.}
	\label{fig:teaser}
	\vspace{-5mm}
\end{figure}

 In this work, we explore robot navigation in crowds of human groups. Our approach is to learn navigation policies that allow the robot to safely reach its desired goal while minimizing impact to individuals and groups of pedestrians (Fig.~\ref{fig:teaser}). Our learned policy only requires knowledge of group membership during training and is able to generalize without group knowledge during inference. Our contributions include:

\begin{itemize}
    \item A reinforcement learning (RL) algorithm based on proximal policy optimization (PPO)\cite{ScWoDhRaKl17} that combines robot navigation performance and group-aware social norms for learning a robust  policy;
    \item A novel reward function that uses the convex hull of a group as the group space to minimize impact to pedestrian groups and improve navigation performance;
    \item Software extensions to the CrowdNav simulation environment~\cite{Chen2019CrowdRobotIC} to support social navigation research; and
    \item Experimental results that demonstrate the efficacy of our learned policy with respect to robot navigation performance, human navigation performance, and maintenance of social norms.  
\end{itemize}

\section{Background and Related Work}
The goal of human-aware robot navigation is to enable robots to move safely, efficiently, and socially appropriately in natural human environments. 
To achieve safe and efficient navigation, prior research has investigated reactive methods for motion planning \cite{van2011reciprocal,luo2018porca} and considered modeling pedestrian intention \cite{katyal2020intent}.  
To realize social appropriateness, previous works have explored learning from human data \cite{robicquet2016learning,luber2012socially,shiomi2014towards,sebastian2017socially,doi:10.1177/0278364915619772,pellegrini2009you} and used handcrafted rules as planning constraints \cite{bera2017sociosense,johnson2018socially,kirby2009companion,truong2016dynamic}.
One notable approach is the Social Force Model ~\cite{helbing1995social}, which is based on the proxemics theory~\cite{hall1966hidden} and attempts to model pedestrian social motion using a combination of attractive and repulsive forces.
This approach has been adapted and extended for crowd simulation \cite{farina2017walking} and robot navigation \cite{ferrer2014proactive,ratsamee2013human}.

While prior research on human-aware robot navigation has mostly regarded humans as individual agents, increasing efforts have considered how mobile robots should interact with human social groups. 
We consider human social groups as two categories---\emph{static} and \emph{dynamic} social groups. Static social groups are commonly seen in social events (e.g., a cocktail party), where people gather together in small groups for conversation. Dynamic social groups are groups of people walking together, and possibly engaging in conversations during walking, toward shared destinations. 
Previous research has investigated how to enable robots to recognize \cite{swofford2019improving} and approach \cite{kato2015may,gomez2014fast} static, standing social groups, while taking account of the size and formation of the groups.

Though the detection and modeling of dynamic social groups present additional challenges when compared to static groups, they are critical in enabling socially appropriate robot navigation in human crowds. 
Prior research has explored methods to capture intra-group coherence (e.g.,~\cite{taylor2020robot, luber2013multi}) and inter-group differences (e.g.,~\cite{shao2014scene}) in dynamic groups. 
For instance, salient turn signals in groups of humans that share the same navigation goals can be used to enhance trajectory prediction and subsequently improve the social-awareness in robot  planning~\cite{unhelkar2015human}. 
However, it has been shown that group properties may be different in static and dynamic settings. For example, the o-space formation commonly observed in static groups is not necessarily apparent in dynamic groups~\cite{yang:2019}. To address such a difference, a set of dynamic constraints for o-space based on the walking direction and group cohesion was proposed for effective robot navigation~\cite{yang:2019}. 
Additionally, dynamic groups also bear unique properties that mobile robots may take advantage of during their navigation in crowded environments. As an example, robots may ``group surf" human groups by following their movement flows~\cite{du2019group}.

Methodologically, modern machine learning techniques have fueled the advances of human-aware robot navigation. 
In particular, Recurrent Neural Networks (RNNs) and Generative Adversarial Networks (GANs) have been shown to be able to accurately predict human motion for individuals \cite{alahi2016social,gupta2018social} and groups \cite{bisagno2018group}. 
However, RNN and GAN based methods only predict human motion and do not generate navigation policies for robots. 
RL approaches are increasingly used for learning navigation policies. 
For example, prior works have leveraged inverse RL to imitate humans and realize socially appropriate movements \cite{okal2016learning,doi:10.1177/0278364915619772}. 
Deep Reinforcement Learning (DRL) has also been used for robot navigation \cite{tai2017virtual, chen2019relational}; in particular, attention-based DRL has been demonstrated to capture human-human and human-robot interactions in crowded environments~\cite{Chen2019CrowdRobotIC,chen2017decentralized}. 

Different from these prior works, we explicitly include group modeling, rather than a simple consideration of pairwise interactions between individuals in a crowd~\cite{chen2019relational}. 
In addition, our approach uses a more compact representation of group space by computing a polygon based on the convex hull of the pedestrians instead of the F-formation as in prior work \cite{yang:2019}. 
Moreover, our approach considers a range of metrics including social compliance and pedestrian and robot performance; our metrics expand on the common measures used in evaluating socially aware robot navigation \cite{gao2021evaluation}.
Finally, our work uses violations of the group space as a reward term to learn socially appropriate movements around groups of pedestrians.

\section{Preliminaries}

\subsection{Problem Formulation}

Our objective is to learn a controller that allows a robot to navigate to a desired goal while maintaining social norms and avoiding collisions with groups of pedestrians.  We formulate our approach using RL to learn a policy to meet the objectives stated above.  In this form of a Markov decision making process, the robot uses observations to generate a state vector, $\textbf{S}$, and chooses an action, $\textbf{A}$, that maximizes expectation of the future reward, $\textbf{R}$.  

\subsection{State and Action Space}
The state space, $\textbf{S}$, consists of observable state information for each pedestrian $i$, represented as $\textbf{Ped}_i$ as well as internal state of the robot represented as $\textbf{Rob}$ as described by Eq.~\ref{eq:state_def}.  Here, $p^{p/r}_x$ and $p^{p/r}_y$ are the position $x$ and $y$ coordinates for the robot and pedestrian, respectively, $v^{p/r}_x$ and $v^{p/r}_y$ are the velocity $x$ and $y$ coordinates for the robot and pedestrian, $rad^{p/r}$ is the radius of the pedestrian or the robot, $g_x$ and $g_y$ represent the $x$ and $y$ goal positions, $v\_pref$ is the preferred velocity and $theta$ is the turn angle.

\begin{equation}
\begin{aligned}
\label{eq:state_def}
	\textbf{Ped}_i &= [p^p_x, p^p_y, v^p_x, v^p_y, rad^p], \\
	\textbf{Rob} &= [p^r_x, p^r_y, v^r_x, v^r_y, rad^r, g_x, g_y, v\_pref, theta], \\
	\textbf{S}_i &= [\textbf{Ped}_i, \textbf{Rob}]
\end{aligned}{}
\end{equation}

In our simulation, we assume a robot with holonomic kinematics that receives $v^r_x$ and $v^r_y$ commands.  The action space is discretized into 5 speeds ranging from 0.2 to 1.0 m/s and 16 rotations ranging from 0 to 2$\pi$ plus a stop command resulting in 81 possible actions.

\subsection{CrowdNav Simulation Environment}

We leverage the CrowdNav simulation environment~\cite{Chen2019CrowdRobotIC} for training and evaluation. This environment provides a simulation framework that allows us to model scenes of pedestrians and robots interacting while reaching their target goals.  We extend this framework by allowing groups of pedestrians to be instantiated with similar starting and end goals.  We further leverage our group-aware social force model described in the next section to model the motion of the groups of pedestrians as they interact with other pedestrians and the robot.  

\subsection{Group-Aware Social Force Model}\label{sec:sf_model}
We use an extended Social Force Model (SFM) \cite{moussaid2010walking} to simulate pedestrian motion in dynamic social groups. 
In the extended SFM, each individual's motion, as defined in Eq. \ref{eq:sfm_def}, is driven by a combination of an attractive force \(\vv{f_i}^{des}\) that drives them to a desired goal, the obstacle repulsive forces \(\vv{f_i}^{obs}\), the sum of social repulsive forces from other agents \(\sum_j\vv{f}_{ij}^{social}\), and a new group term \(\vv{f}_i^{group}\) defined by Eq. \ref{eq:group_term_def}.
\begin{align}
\label{eq:sfm_def}
  \frac{d \vv{v_i}}{d t} =& \vv{f}_i^{des} + \vv{f}_i^{obs} + \sum_j\vv{f}_{ij}^{social} + \vv{f}_i^{group}
\end{align}
The group term is defined as the sum of the attractive forces between group members \(\vv{f}_i^{att}\), the repulsive force between group members \(\vv{f}_i^{rep}\), and a gaze force \(\vv{f}_i^{gaze}\) that steers the agents to keep the center of mass of the social group within their vision field to simulate in-group social interactions:
\begin{equation}
\label{eq:group_term_def}
  \vv{f}_i^{group} = \vv{f}_i^{att} + \vv{f}_i^{rep} + \vv{f}_i^{gaze} 
\end{equation}

We developed a custom Python implementation of the extended Social Force Model~\footnote{https://github.com/yuxiang-gao/PySocialForce} for the CrowdNav environment, following the implementation of PEDSIM C++ library~\cite{gloor2003pedsim} and the ROS implementation by Vasquez et al.~\cite{vasquez2014inverse}.


\section{Approach}\label{sec:approach}

To evaluate our group-aware policy, we extend the existing CrowdNav simulation environment~\cite{Chen2019CrowdRobotIC} to represent pedestrian motion in groups. We accomplish this by stochastically sampling the number of group members per episode using a Poisson distribution ($\lambda=1.2$)~\cite{coleman1961equilibrium} and then randomly assigning pedestrians to the group.  Each pedestrian within a group has similar start and goal positions.  Based on this sampling strategy, the average number of groups and group size for five pedestrians are 2.5 and 1.96, respectively.  For ten pedestrians, the number of groups and group size increase to 4.9 and 2.0, respectively.  

\begin{figure*}[t] 
	\centering
    \includegraphics[width=1.75\columnwidth]{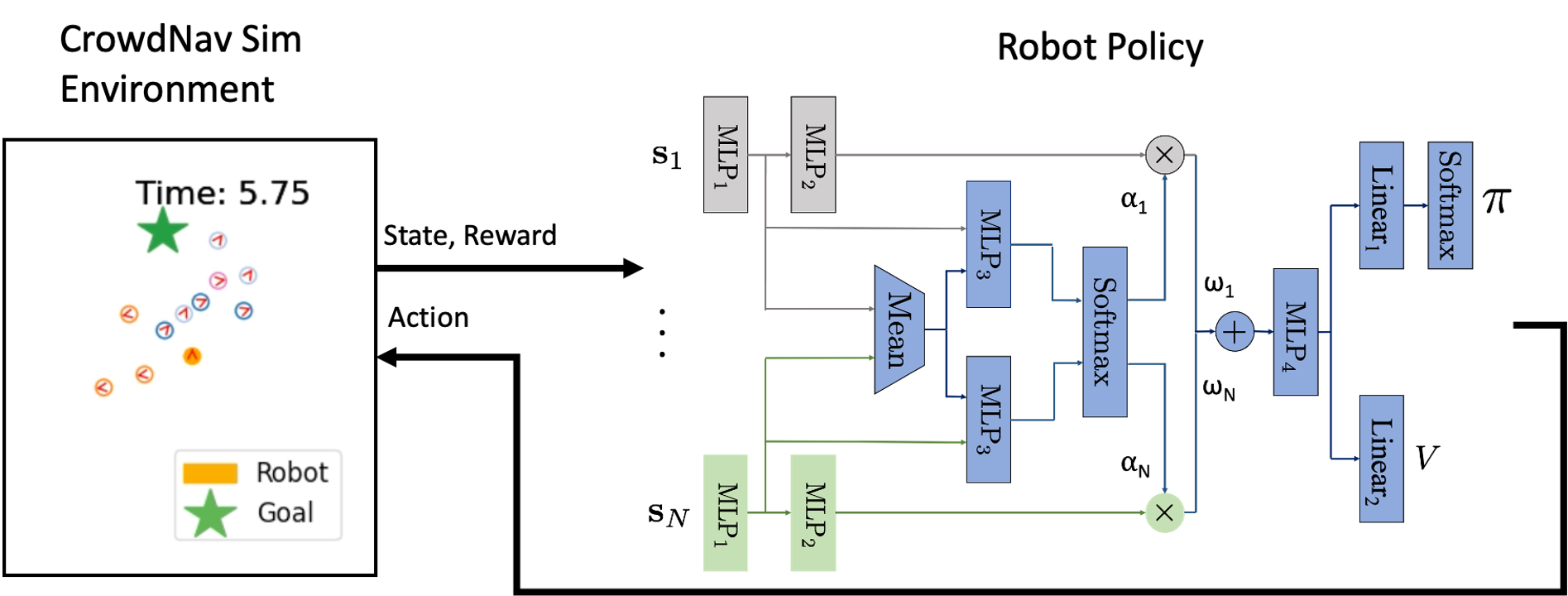}
    \caption{On the left is the CrowdNav Simulation~\cite{Chen2019CrowdRobotIC} that provides pedestrian, robot state and reward information to the policy.  The right figure represents the network architecture used for our attention-based, actor-critic policy.  The pedestrian and robot state vectors are concatenated to represent a pairwise combined state vector; the output of the network are the policy $\pi$ over potential actions and value $V$ of the current state. The gray and green blocks indicate features from individual pedestrians.  The blue blocks indicate aggregate features across pedestrians.  The argmax of the policy is chosen as the action, which is sent to the CrowdNav Simulation to control the robot.}
	\label{fig:network}
\end{figure*}

\subsection{Policy based on Convex Hull of Group}

To train the policy, we use a multi-term reward function that encourages the robot to reach its goal while maintaining social norms and avoiding collisions with groups of pedestrians.  In particular, we focus on social norms that minimize discomfort to individuals and discourage intersections with a group of pedestrians.

Our reward function is given by Eq.~\ref{eq:reward_def}, where $d_{\text{goal}}$ is the distance from the robot to the goal, $d_{\text{coll.}}=0.6$ is the distance between the centers of entities which a collision is considered to have occurred, $d_i$ is the distance between the robot and pedestrian $i$, $d_{\text{disc.}}=d_{\text{coll.}}+0.2$ is the minimum ``comfortable" distance between a robot and a pedestrian~\cite{Chen2019CrowdRobotIC}, and $d^g_j$ is the distance from the robot to the edge of the convex hull surrounding group $j$ and $\mathbb{1}$ is the Indicator function:

\begin{equation}
\begin{aligned}
\label{eq:reward_def}
R(t) = 
& C_{\text{prog.}}(d_{\text{goal}}(t-1) - d_{\text{goal}}(t)) \\
& + C_{\text{goal}}\mathbb{1}(d_{\text{goal}}(t) < d_{\text{coll.}})\\
& - C_{\text{disc.}}\sum_i (d_{\text{disc.}} - d_i(t))\mathbb{1}(d_{\text{coll.}} \le d_i(t) \le d_{\text{disc.}}) \\
& - C_{\text{coll.}} \sum_i {\mathbb{1}(d_i(t) < d_{\text{coll.}})} \\
& - C_{\text{group}} \sum_j {\mathbb{1}(d^g_j(t) < d_{\text{coll.}})}. \\
\end{aligned}{}
\end{equation}

\noindent The multiple objectives are weighted via the following constants: $C_{\text{prog.}} = 0.1$, $C_{\text{goal}} = 1.0$, $C_{\text{disc.}} = 0.5$, $C_{\text{coll.}}=0.25$, and $C_{\text{group}}=0.25$.  The first term encourages the robot to progress toward the goal, allowing us to remove the initial imitation learning phase in \cite{Chen2019CrowdRobotIC}.  The second, third, and fourth terms encourage the robot to reach the goal, avoid close encounters with pedestrians, and avoid collisions, respectively.  The last term encourages the robot to follow group social norms by penalizing any group space violation.

To determine the $d^g_j$ terms, we first compute a polygon representing the convex hull of the positions of all members of the pedestrian group. We then calculate the minimum distance between the robot and the polygon and penalize the robot for intruding into this space. Note that group membership information is only needed during training and is not needed for evaluation or deployment of the policy.

\subsection{Neural Network Architecture}
Our overall network architecture is depicted in Fig.~\ref{fig:network}.  As in \cite{Chen2019CrowdRobotIC}, we used an attention-based  architecture to represent navigation policies.  For each pedestrian, a vector representing the pedestrian was first concatenated with a vector representing the robot and passed through the first multi-layer perceptron (MLP) in the network ($\text{MLP}_1$).  The resulting feature vector was concatenated with the mean value of the outputs of $\text{MLP}_1$ for all pedestrians and used to compute an attention score $\alpha_i$ for each pedestrian via $\text{MLP}_3$.  The output of $\text{MLP}_1$ was also passed through $\text{MLP}_2$ to generate a separate ``robot-pedestrian interaction vector" that was then multiplied by $\alpha_i$ to generate a weighted feature vector $\omega_i$ for each pedestrian.   The $\omega_i$ were summed for all pedestrians and the result was passed through $\text{MLP}_4$, leading to separate policy and value heads. The filter parameters for the network architecture are described in Table~\ref{table:parameters}.  In summary, our architecture matched that of \cite{Chen2019CrowdRobotIC} without the interaction module and with 1) a softmax layer being added to produce a categorical policy output and 2) a single fully-connected layer with 100 neurons connecting to a scalar value head.  

\subsection{PPO Algorithm}
Our agents were trained using proximal policy optimization (PPO; \cite{ScWoDhRaKl17}), a leading model-free, actor-critic approach.  Hyperparameters were chosen to mimic those used for Atari in \cite{ScWoDhRaKl17}, with the exceptions of shorter windows (16 steps) and more windows per batch (64).  This change was made to accommodate the shorter episodes of CrowdNav while maintaining the number of experiences per batch.

We used the Adam optimizer~\cite{DBLP:journals/corr/KingmaB14} with learning rate set to \num{2.5e-4} and epsilon set to \num{1.0e-5}.  In the RL policy, the discount factor, $\gamma$, was set to 0.99 and the credit assignment variable, $\lambda$, was set to 0.95.  We trained our policy for 7000 iterations yielding approximately 4.8M steps and reaching a maximum reward of approximately 1.7 based on the reward definition described in Eq.~\ref{eq:reward_def}.

\begin{table}[]
  \centering
   \setlength{\tabcolsep}{1.2mm}
   \renewcommand{\arraystretch}{1.1}
  \begin{tabular}{|c|c|}
  \hline
  \textbf{Network Layer} &  \textbf{Output Features / Activation}\\
  
  \hline
  
  \({MLP}_1\)  & 150, ReLU, 100, ReLU \\
  \hline
  \({MLP}_2\) & 100, ReLU, 50 \\
  \hline
  \({MLP}_3\) & 100, ReLU, 100, ReLU, 1 \\
  \hline
  \({MLP}_4\) &  150, ReLU, 100, ReLU, 100, ReLU\\
  \hline
  \({Linear}_1\) & 81 \\
  \hline
  \({Linear}_2\) & 1 \\
  \hline

   \end{tabular}
  \caption{Network Layer Filter Parameters}
  \label{table:parameters}
  \vspace{-5mm}
\end{table}

\section{Experimental Evaluation}

\subsection{Experimental Setup}
The goal of our experiment is to assess the efficacy of our group-aware navigation policy. 
Our experiments involved four settings determined by two factors: the number of pedestrians and the number of groups. We explored both 5-person and 10-person settings as well as a single group and a stochastic number of groups as described in Sec.~\ref{sec:approach}.

We used the Circle Crossing scenario where groups of pedestrians started and ended around the perimeter of a circle (radius = 4 m) during training and evaluation.  The robot's starting and end positions were set to ensure the robot goes through the center of the circle and interacts with the pedestrian groups. 
We evaluated our trained policy on 250 trials with randomly initialized starting and ending pedestrian positions for the four experimental settings.   

\subsection{Metrics}
Our evaluation was focused on 1) robot navigation performance, 2) pedestrian navigation performance, and 3) social compliance. 
For robot navigation performance, our metrics represent the quality of the robot's ability to navigate to the goal quickly without collision:  

\begin{itemize}
\item \textbf{Successes}: Number of trials the robot reached the goal.
\item \textbf{Collisions}: Number of trials a collision occured. 
\item \textbf{Timeouts}: Number of trials the robot did not reach the goal within the allotted time (25 seconds).
\item \textbf{Time to Goal}: Average of the number of seconds the robot needed to reach the goal for all trials.
\item \textbf{Mean Robot Velocity}: The average velocity of the robot at each time step during all trials.
\end{itemize}

To assess pedestrian performance, we measured the impact of the robot's behavior on the desired pedestrian motion:

\begin{itemize}
\item \textbf{Mean Pedestrian Velocity}: The average velocity of the pedestrians during the trials.
\item \textbf{Mean Pedestrian Angle}: The average angular deviation between the pedestrian's observed motion and the direct vector to the pedestrian's goal. This metric seeks to measure the disturbance from the optimal trajectory to the goal caused by the robot's policy.
\end{itemize}

Finally, to assess social norms, we quantified how the robot maintained social distance among individual pedestrians and limited intersections with groups of pedestrians: 

\begin{itemize}
\item \textbf{Group Intersections}: The number of groups intersections by the robot that occurred during the trials.
\item \textbf{Individual Discomfort}: The mean distance between the robot and the pedestrians aggregated over all pedestrians when the robot violates the discomfort threshold (0.2 m). 

\item \textbf{Pedestrian Social Force}: The mean social force applied to pedestrian $i$.  This is equal to the sum of the forces applied to pedestrian $i$ from the other pedestrians and the robot, $j$ as described in Sec.~\ref{sec:sf_model} (i.e., $\sum_j\vv{f}_{ij}$). This metric captures how the robot's motion may impact directly or indirectly human pedestrians' motions.
\item \textbf{Robot Social Force}: The mean social force applied to the robot, $r$ from other pedestrians, $j$ as described in Sec.~\ref{sec:sf_model} (i.e., $\sum_j\vv{f}_{rj}$). This metric captures how the robot's motion may be impacted by human pedestrians.
\end{itemize}

\begin{table*}[t!]
  \centering
  \setlength{\tabcolsep}{1.2mm}
  \renewcommand{\arraystretch}{1.1}
  \begin{tabular}{|l|l|l|l|l|l|l|l|l|l|l|l|l|l|}
  \hline
  \multirow{3}{*}{\textbf{Method}} &  
  \multirow{3}{0.9cm}{\textbf{\# Groups}} & 
  \multirow{3}{0.6cm}{\textbf{\# Peds.}} & 
  \multirow{3}{0.7cm}{\textbf{Succ. $\uparrow$}} & 
  \multirow{3}{0.5cm}{\textbf{Ped. Coll. $\downarrow$}} & 
  \multirow{3}{0.4cm}{\textbf{TO $\downarrow$}} & 
  \multirow{3}{0.8cm}{\textbf{Mean Time (s) $\downarrow$}} & 
  \multirow{3}{1.3cm}{\textbf{Mean Robot Vel. (m/s) $\uparrow$}} & 
  \multirow{3}{1.1cm}{\textbf{Mean Ped. Vel. (m/s) $\uparrow$}} & 
  \multirow{3}{1.4cm}{\textbf{Mean Ped. Ang. (\degree) $\downarrow$}} &
  \multirow{3}{0.8cm}{\textbf{Grp. Inters. $\downarrow$}} & 
  \multirow{3}{1.4cm}{\textbf{Ind. Discomfort $\downarrow$}} & 
  \multirow{3}{1.0cm}{\textbf{Ped. Social Force $\downarrow$}} & 
  \multirow{3}{1.0cm}{\textbf{Robot Social Force $\downarrow$}} \\ 
  & & & & & & & & & & & & & \cr   
& & & & & & & & & & & & & \cr  
  \hline
  \hline
SARL & 1 & 5 & \textbf{237} & 11 & \textbf{2} & \textbf{8.24}*  & 0.962 & 1.170* & 3.76 & 143 & 3.10* & 0.375* & 0.523 \\

SocialNCE (w/o CL) & 1 & 5 & 187 & 60 & 3 & 10.54*  & 0.777* & 1.139* & 3.61 & 64 & 3.06* & 0.391* & 0.545* \\

SocialNCE & 1 & 5 & 213 & 31 & 6 & 10.87* & 0.796*  & 1.166*  & \textbf{3.43} & 50 & 2.72* & \textbf{0.349} & \textbf{0.473}  \\

Group-Aware (ours) & 1 & 5 & 236 & \textbf{9} & 5 & 8.92 &  \textbf{0.964} &  \textbf{1.183} &  3.59 & \textbf{15} & \textbf{1.29} & 0.351 & 0.482 \\
\hline
\hline

SARL & 2.548 & 5 & 238 & 12 & 0 & \textbf{8.23}* & \textbf{0.964} & 1.136  & 5.99* & 151 & 2.87 & 0.522* & 0.716*  \\

SocialNCE (w/o CL) & 2.548 & 5 & 174 & 76 & 0 & 9.73* &  0.797* & 1.107* & 5.71 & 50 & 3.91* & 0.539* & 0.648  \\

SocialNCE & 2.548 & 5 & 206 & 44 & 0 & 9.68* &  0.817* & 1.125*  & \textbf{5.25}* & 56 & \textbf{2.36} & 0.489 & \textbf{0.582}*  \\

Group-Aware (ours) & 2.548 & 5 & \textbf{242} & \textbf{8} & 0 & 8.81 & 0.961 &  \textbf{1.146} & 5.59 & \textbf{22} & 2.63 & \textbf{0.485} & 0.657 \\
\hline
\hline
SARL & 1 & 10 & 222 & 23 & 5 & \textbf{8.59}*   & 0.955&  1.161*& 4.11& 176 & 4.20* & 0.395* & 0.707* \\

SocialNCE (w/o CL) & 1 & 10 & 127 & 121 & \textbf{2} & 10.92* &  0.733* & 1.090*  & 4.17* & 85 & 3.30 & 0.465* & 0.944*  \\

SocialNCE & 1 & 10 & 167 & 76 & 7 & 11.32* &  0.728* & 1.131*  & 4.02 & 94 & 4.29* & 0.413* & 0.795*  \\

Group-Aware (ours) & 1 & 10 & \textbf{232} & \textbf{14} & 4 & 9.87 & \textbf{0.956} & \textbf{1.174} & \textbf{3.93} & \textbf{29} & \textbf{2.31} & \textbf{0.366} & \textbf{0.597}\\
\hline
\hline

SARL & 4.884 & 10 & 239 & 10 & 1 & \textbf{8.72}* &  \textbf{0.960}* &  1.089* &  8.09* & 258 & 4.94* & 0.681* & 0.964*\\

SocialNCE (w/o CL) & 4.884 & 10 & 141 & 109 & \textbf{0} & 10.20 &  0.693* & 1.022*  & 8.18* & 95 & 5.71* & 0.764* & 0.986*  \\

SocialNCE & 4.884& 10 & 186 & 64 & \textbf{0} & 9.89* &  0.772* & 1.074 & 7.81* & 156 & 5.71* & 0.684* & 0.928*  \\

Group-Aware (ours) & 4.884 & 10 & \textbf{241} & \textbf{9} & \textbf{0} & 10.21 &  0.918&  \textbf{1.108}* & \textbf{7.07}& \textbf{20} & \textbf{2.29} & \textbf{0.599} & \textbf{0.849} \\
\hline

  \end{tabular}
  \caption{
  This table summarizes our results across 5 and 10 pedestrians. The best values are marked with bold texts. Arrows represent whether a higher or lower number is desirable. Asterisks indicate statistically significant results ($p<.05$) when compared to our group-aware policy.  We show that our policy is able to achieve comparable or better robot navigation performance while allowing pedestrians to achieve faster velocities. In addition, we show our policy leads to more socially compliant navigation indicated by significantly fewer instances of group intersection while reducing individual discomfort.
    }
  \label{table:r_and_p_results_summary}
\end{table*}

\begin{figure*}[t!] 
	\centering
    \includegraphics[width=1.0\textwidth]{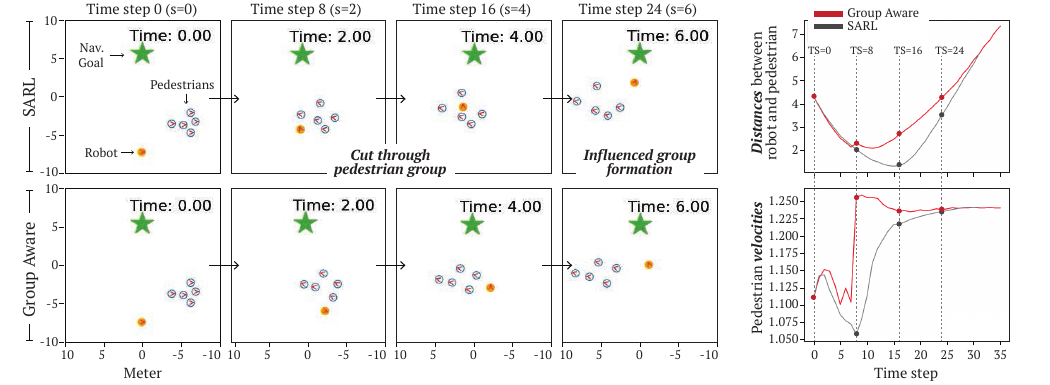}
    \caption{The left figure shows a representative example of the robot navigating through the crowd of pedestrians over time using both the SARL and the group-aware policy.  
    The SARL policy chooses actions that intersect through the group of pedestrians and influences the group formation, while our group-aware policy chooses actions that move around the group with minimal disturbance. 
    The right figure shows the average distance between the pedestrian and the robot (top) and the average pedestrian velocity (bottom) over time.  Here, we show the group-aware policy results in increased distance to the pedestrians while allowing the pedestrians to maintain faster speeds. }
	\label{fig:case-study}
	\vspace{-5mm}
\end{figure*}

\subsection{Results}
We conducted independent two-tailed t-tests to compare our group-aware to the following baseline policies. 
\begin{itemize}
\item \textbf{SARL}: Crowd aware RL policy that uses an attention mechanism to model human robot interaction to encourage socially compliant navigation~\cite{Chen2019CrowdRobotIC}.
\item \textbf{SocialNCE (w/o CL)}: Social NCE policy without contrastive loss. This policy uses behavior cloning using a trained SARL policy as the expert \cite{liu2020snce}.
\item \textbf{SocialNCE}: Social NCE policy. This policy uses a contrastive loss function to represent `negative' examples such as collisions to improve social compliance \cite{liu2020snce}.
\end{itemize}

Table~\ref{table:r_and_p_results_summary} summarizes the robot and pedestrian navigation performance as well as the social compliance results. Generally, the group-aware policy led to higher number of successful trials, while allowing the pedestrians to travel at faster speeds with less disturbance towards the goal. The group-aware policy led to a significant fewer number of instances where the robot navigated through a group. The group-aware policy also resulted in an overall less individual discomfort. Finally, we observe that our policy improved the overall social forces applied to the pedestrians and robot.


We note that the robot with the group-aware policy took longer to reach the goal compared to the SARL policy.  However, we do not see significant differences in robot speed between the two approaches. The group-aware robot took a longer path to its goal due to its preference of navigating around groups of pedestrians to avoid group intersections, whereas the SARL robot aimed to reach its goal in spite of cutting in between groups of pedestrians. 
Fig.~\ref{fig:case-study} illustrates an example of such behavior---the SARL policy chose a path that intersects the pedestrian group whereas, in the same scenario, our policy chose a path around the group. The resulted group-aware behavior ultimately enabled greater group cohesion and less disruption, while improving group and individual discomfort. 
We additionally computed the distances between the pedestrians and the robot for both policies (Fig.~\ref{fig:case-study} top-right), as well as the velocities of the pedestrians (Fig.~\ref{fig:case-study} bottom-right), over time. 
During interaction between the robot and the pedestrians, we observe that the distance between the pedestrians and the robot was lower for the SARL policy corresponding to the results reported in Table~\ref{table:r_and_p_results_summary}. Further, we see that the average pedestrian velocity decreased substantially during the times of interaction in the SARL policy; however, we do not see similar decreases in our group-aware policy. 

\subsection{Hardware Demonstration}

To demonstrate the applicability of our group-aware policy to real-world systems, we implemented our policy on the Boston Dynamics Spot robot. The policy was trained entirely in the CrowdNav environment before the demonstration, then evaluated on the Spot robot as a proof of concept. In this demonstration, the robot's start and goal positions were predetermined such that the robot reached a point approximately six meters in front of its start position. The radius and preferred velocity of the robot and all of the pedestrians were approximated ahead of time and set when the robot and pedestrian objects were defined. 


During the demonstration, the pose of the robot and each pedestrian was determined using an OptiTrack motion capture system, then the velocity of each agent was approximated appropriately. The observable state of the pedestrians and the full state of the robot was used as input to our group-aware policy. The policy then predicted the best action as $v^r_x$ and $v^r_y$, which was mapped to a ROS command used to control the robot. We decreased the speed in both the $x$ and $y$ direction by a factor of two to ensure that the robot moved at an appropriate speed to compensate for hardware limitations. 

As illustrated in Fig. \ref{fig:teaser}, we show that the policy worked on a real-world demonstration by having pedestrians move around the robot in different formations in a way that interfered with the path of the robot. To demonstrate the effectiveness of the policy, the pedestrians followed paths that varied from the circle crossing movements seen by the simulated robot during training. Videos of these demonstrations can be found in the supplementary attachment.

\section{Discussion}
Towards achieving socially appropriate robot navigation in human environments, this paper explores group-aware behaviors that respect pedestrian group formations and trajectories, while minimally sacrificing robot navigation performance. 
Our results show that the learned policy is able to achieve a higher number of successful trials in which the robot reached the goal, fewer collisions, and less impact to the pedestrian's motion towards their goal. In addition, we show that our learned policy not only reduced the number of group violations (e.g., intersecting the group) but also decreased the individual discomfort and social forces applied to the pedestrians and robot. 
Our approach, however, resulted in an increase of the robot's total time to goal compared to the SARL baseline that did not consider group formation. 
This increase of time was expected as the robot sought to move around groups as opposed to navigating through them (Fig. \ref{fig:case-study}). However, our results show that even though the total time to goal increased, the average velocity of the robot was mostly unaffected by the group-aware policy.

Our exploration indicates several directions of future research. 
First, we would like to determine how well our policy reflects actual human motion through groups of pedestrians and investigate whether we can bootstrap our learned policy with imitation learning using observations of humans navigating around groups of pedestrians.  
Second, we plan to investigate different representations of group space beyond the convex hull approach described in this paper. Considering additional parameters, such as social interaction during movement, the specific formation of the group, and environmental cues (e.g., social space), may contribute to learning more socially compliant navigation policies. 
Third, beyond those discussed in this paper, there are other factors to consider for socially appropriate behavior, such as how to pass and follow human groups.
Finally, while learning a single policy works for a limited number of environments, choosing amongst a library of policies depending on the density of pedestrians, the layout of the environment, and the local culture can lead to better navigation performance once deployed. 
For example, respecting human grouping may not always be possible (e.g., in a narrow corridor). It is therefore important for a robot to selectively choose a context suitable policy in order to achieve efficient, safe, and socially appropriate navigation in crowded human environments. 

\section*{Acknowledgments}
This work was supported by the Johns Hopkins University Institute for Assured Autonomy.
\balance
\bibliographystyle{IEEEtran}
\bibliography{references2,references,references_icra2020,bibliography1}

\end{document}